%% file: acl_latex.tex
\documentclass[11pt]{article}
\usepackage[preprint]{acl}     
\usepackage{times}
\usepackage{latexsym}
\usepackage[T1]{fontenc}
\usepackage[utf8]{inputenc}
\usepackage{microtype}
\usepackage{inconsolata}

\usepackage{graphicx}
\usepackage{booktabs}
\usepackage{amsmath,amssymb,amsfonts,nicefrac,mathtools}
\usepackage{xcolor,colortbl}
\usepackage{makecell,tabularx,multirow,enumitem,listings}
\usepackage{url}

\usepackage[capitalize,noabbrev]{cleveref}
\usepackage{subcaption}
\usepackage{algorithm}
\usepackage{algorithmic}
\usepackage{tikz}
\usetikzlibrary{shapes,arrows,positioning,fit,calc}

\newcommand{\benchname}{ShoppingReasoningBench}
\newcommand{\benchnameFull}{\benchname{}-Full}

\newcommand{\benchnameHard}{\benchname{}-Hard}
\newcommand{\passrate}{\text{WPR}}
\definecolor{csrbblue}{RGB}{51,102,170}
\definecolor{csrborange}{RGB}{204,102,0}
\definecolor{csrbgray}{RGB}{128,128,128}
\definecolor{lightgray}{RGB}{240,240,240}

\newif\ifshowcomments
\showcommentsfalse

\title{Shopping Reasoning Bench: An Expert-Authored Benchmark for Multi-Turn Conversational Shopping Assistants}
\author{
  \textbf{Shuxian Fan}\thanks{\ Equal contribution.}\quad
  \textbf{Seonwoo Min}\footnotemark[1]\quad
  \textbf{Youna Hu}\quad
  \textbf{Botao Xia}\quad
  \textbf{Jayakrishnan Unnikrishnan}\\
  \textbf{Rowan Musselmann}\quad
  \textbf{Yifan Gao}\quad
  \textbf{Qingyu Yin}\quad
  \textbf{Priyanka Nigam}\quad
  \textbf{Bing Yin}\\[0.5em]
  Amazon\\
  \texttt{\{fansx, seonwoom, ynhu, xiabota, jayunn, saramuss, yifangao, qingyy, }\\ \texttt{nigamp, alexbyin\}@amazon.com}
}

\begin{document}
\maketitle

\input{sections/abstract}

\input{sections/introduction}
\input{sections/related_work}

\input{sections/shopping_taxonomy}

\input{sections/rubric_evaluation}

\input{sections/results}

\input{sections/discussion}

\input{sections/conclusion}

\input{sections/limitations}
\input{sections/ethics}

\section*{Acknowledgments}
We thank our domain expert annotation team---Rowan Musselmann, Elizabeth Gongliewski, Jastine Sanchez, Kenneth Young, Laura Santana, Tom Knee, and others---for their meticulous work in constructing the evaluation rubrics and expert reasoning traces that underpin this benchmark.

\bibliography{references}

\appendix
\input{appendices/appendix_taxonomy}

\input{appendices/appendix_annotation}
\input{appendices/appendix_model_details}

\input{appendices/appendix_judge_prompts}
\input{appendices/appendix_results}
\input{appendices/appendix_benchmark_variants}
\input{appendices/appendix_data_examples}

\end{document}

%% file: sections/abstract.tex
\begin{abstract}
Conversational shopping assistants now serve hundreds of millions of customers, yet no existing benchmark jointly evaluates the open-ended multi-turn reasoning, domain expertise, and criterion-level quality that real shopping conversations demand. Shopping reasoning is unique among language model applications. Unlike factual question answering or verifiable code generation, it requires balancing subjective preferences, budget constraints, and cross-product trade-offs across multi-turn dialogue, capabilities absent from previous e-commerce and general-purpose benchmarks. We introduce the \textbf{Shopping Reasoning Bench}, an expert-authored benchmark of 525 missions (232 single-turn, 293 multi-turn) with 10{,}863 importance-weighted binary rubrics authored by retail domain experts. These criteria are organized under a taxonomy of five reasoning categories and fifteen subcategories covering diverse demands such as preference refinement, trade-off analysis, and compatibility assessment. An evaluation of nine models across three families (GPT, Claude, Gemini) shows that pass rates reach only 57--77\% overall. On multi-turn missions, all models score 13--29 points lower on optional above-and-beyond criteria than on required ones, and performance degrades 4--18 points as conversations progress. These gaps show that current models handle basic shopping assistance but fall short of expert-level advice, making \textbf{Shopping Reasoning Bench} a challenging testbed for future shopping assistant development.
\end{abstract}

%% file: sections/introduction.tex
\section{Introduction}
\label{sec:introduction}

Suppose a customer asks an AI shopping assistant to recommend trail running shoes sturdy enough for backpacking. The assistant lists five popular models but never warns that cushioned soles lose stability under load and never suggests sizing up to accommodate foot swelling on long hikes. The response answers the question, but a retail expert would call it shallow.

Conversational shopping assistants have reached consumer scale: Amazon Rufus serves over 300 million customers~\citep{amazon2026q4earnings}, and major search platforms including Perplexity have integrated AI-powered shopping features~\citep{perplexity2024shopping}. Evaluating these assistants is harder than evaluating many conventional language model applications. A good shopping response must balance subjective preferences, budget constraints, and product trade-offs across a multi-turn conversation with evolving user intent, all while drawing on product-domain expertise.

\paragraph{The benchmark gap.}
A useful benchmark for this setting must be \textbf{(i) grounded in domain expertise} to capture product-specific knowledge that crowd annotators often lack; \textbf{(ii) rubric-verifiable at the criterion level} to resolve the fine capability differences that coarse aggregated scores obscure; and \textbf{(iii) open-ended and multi-turn} to reflect the iterative nature of real shopping conversations. No existing shopping benchmark meets all three requirements (\S\ref{sec:relwork}). \benchname{} is, to our knowledge, the first to jointly satisfy them: it pairs domain-expert-authored rubric criteria with a shopping reasoning taxonomy and evaluates models across multi-turn shopping missions.

\paragraph{Why shopping reasoning?}
The complexity above is not incidental: pre-purchase shopping is a form of \emph{practical reasoning}~\citep{bratman1987intention}, deliberation whose output is a decision to act, not a truth to verify. Answering ``What are the best trail runners that are supportive enough to use for backpacking too?'' requires decomposing the customer's constraints, identifying candidate products, applying domain expertise to evaluate each against those constraints, and synthesizing a recommendation. No single step is retrievable; each depends on the novel intersection of the customer's needs with product-specific knowledge. Existing reasoning benchmarks, mathematical~\citep{cobbe2021gsm8k,hendrycks2021math}, logical and scientific~\citep{suzgun2023bbh,rein2023gpqa}, or code~\citep{chen2021humaneval,jimenez2024swebench}, span a wide difficulty spectrum yet share a defining property: a unique verifiable answer exists. Shopping reasoning has no ground-truth answer, only better and worse deliberation, precisely the gap \benchname{}'s expert-authored rubrics are designed to measure.


\paragraph{Contributions.}
\begin{itemize}
\setlength{\itemsep}{2pt}
  \item \textbf{First taxonomy of pre-purchase shopping reasoning.} Five categories and fifteen subcategories grounded in expert-annotated turns. These capture shopping-specific reasoning patterns that prior shopping-intent~\citep{sondhi2018taxonomy} and product-QA~\citep{yang2024bespoke} taxonomies don't address (\S\ref{sec:taxonomy}). 
\item \textbf{Expert-authored multi-turn shopping dataset.}
232 single-turn queries and 293 multi-turn missions (1{,}764 turns)
authored by retail domain experts across five product families
(\S\ref{sec:taxonomy}).
  \item \textbf{Importance-weighted atomic rubric framework with validated LLM-as-judge.} 10{,}863 binary criteria (85.0\% required) that decompose expert shopping reasoning into independently verifiable pass/fail checks. The LLM judge is validated against expert consensus with per-criterion macro-F1 benchmarked against an inter-expert ceiling (\S\ref{sec:rubric_evaluation}).
  \item \textbf{Empirical study across three model families and capability tiers.} Nine models from the GPT, Claude, and Gemini families, each at frontier, mid, and small tiers. The benchmark separates families, separates tiers within each family, and exposes multi-turn degradation as conversations grow longer (\S\ref{sec:results}).
\end{itemize}
Our benchmark data, judge prompts, and per-model outputs are publicly released at \url{https://huggingface.co/datasets/amazon/ShoppingReasoningBench}.

\paragraph{Headline findings.}
We evaluate nine models across three families and three capability tiers on \benchname{}. First, the benchmark is unsaturated: pass rates range from 57\% to 77\% across the nine models. Second, all models score 13--29 points lower on optional rubrics than on required ones, exposing a persistent gap between basic and expert-level shopping assistance. Third, multi-turn performance degrades 4--18 points over the course of a mission, paralleling the ``lost-in-conversation'' phenomenon~\citep{laban2025lost}.

%% file: sections/related_work.tex
\section{Related Work}
\label{sec:relwork}

\benchname{} draws on shopping-domain benchmarks, expert-authored rubric
benchmarks in other domains, query and intent taxonomies, and multi-turn
LLM evaluation.  Table~\ref{tab:benchmark_comparison} positions \benchname{}
against the most directly comparable shopping benchmarks and against the
rubric-benchmark methodologies from which its evaluation design is
adapted.

\paragraph{Shopping and e-commerce benchmarks.}
Evaluation of conversational shopping assistants has been fragmented across task formulations.  WebShop~\citep{yao2022webshop} benchmarks LLM agents on simulated web navigation, focusing on product selection rather than open-ended reasoning.  Shopping MMLU~\citep{jin2024shoppingmmlu} provides a broad suite of classification-style tasks, but evaluates single-turn closed-form answers. eCeLLM~\citep{peng2024ecellm} constructs instruction-tuning data for e-commerce. ShoppingBench~\citep{wang2025shoppingbench} provides intent-grounded agent tasks against a large product sandbox, measuring end-to-end success rate rather than response quality. EcomEval~\citep{xie2025ecomeval} evaluates shopping assistants across seven languages but does not provide expert-authored rubrics for open-ended scoring. SessionIntentBench~\citep{yang2025sessionintent} models inter-session intention shifts using a hierarchical intention tree, but evaluates with classification metrics.  On the dialogue-dataset side, Wizard of Shopping~\citep{li2025wizardofshopping} and MG-ShopDial~\citep{bernard2023mgshopdial} provide conversational shopping dialogues but lack rubric annotations for automatic criterion-level scoring.

The closest direct competitors are two recent rubric-based shopping benchmarks. ShoppingComp~\citep{tou2025shoppingcomp} introduces an expert-curated single-turn benchmark with rubric-graded product retrieval, report generation, and safety-critical decision-making evaluation. SmartShopBench~\citep{cheng2026chatshopbuddy} introduces a hierarchical two-level evaluation across shopping intent categories, designed to support RL-based agent training.  Both are single-turn and do not organize queries around a published shopping-reasoning taxonomy. \benchname{} differs along three dimensions. First, it adds expert-authored multi-turn missions structured as shopping journeys spanning exploration, comparison, and goal-directed search, alongside single-turn queries that themselves require decomposing customer constraints, identifying candidate products, and weighing trade-offs against domain knowledge. Second, it organizes queries around a published taxonomy of pre-purchase shopping reasoning. Third, its rubric criteria carry importance weights rather than uniform pass/fail.

\paragraph{Expert-authored rubric benchmarks.}
Expert-authored rubric benchmarks have emerged in several domains as general-purpose benchmarks approach saturation. HealthBench~\citep{arora2025healthbench} evaluates models on thousands of multi-turn health conversations scored against rubric criteria written by physicians, validating an LLM judge against physician consensus.  PRBench~\citep{akyurek2025prbench} extends this methodology to finance and law.  ProfBench~\citep{wang2025profbench} covers chemistry, physics, finance, and consulting domains at the PhD and MBA level with response-criterion pairs annotated by domain experts.  Earlier expert-authored benchmarks in science (GPQA~\citep{rein2023gpqa}) and software engineering (SWE-bench~\citep{jimenez2024swebench}) use verifiable answers rather than open-ended rubrics.  \benchname{} extends this lineage to retail, adapting the importance-weighted atomic-criterion protocol and LLM-judge validation design to shopping reasoning.

\paragraph{Query and intent taxonomies.}

Query taxonomies for e-commerce have focused on search-query intent~\citep{sondhi2018taxonomy} or product-QA type~\citep{yang2024bespoke} rather than conversational reasoning; general-purpose dialogue taxonomies like INFINITY-CHAT~\citep{jiang2025infinitychat} are not shopping-specific. None of these resolves the reasoning patterns that distinguish pre-purchase shopping conversations from search or general chat: preference refinement, cross-product trade-off analysis, compatibility assessment, and multi-turn purchase-decision progression. \benchname{}'s taxonomy fills this gap (\S\ref{sec:taxonomy}).


\paragraph{Reasoning in LLM evaluation.}

Existing reasoning benchmarks---mathematical~\citep{cobbe2021gsm8k,hendrycks2021math}, scientific~\citep{suzgun2023bbh,rein2023gpqa}, and code~\citep{chen2021humaneval,jimenez2024swebench}---span a wide difficulty range but share a defining property: a unique correct answer that can be automatically checked. Shopping reasoning fundamentally lacks this property (\S\ref{sec:introduction}); \benchname{} adapts rubric-graded evaluation to this regime, decomposing deliberation into independently verifiable criteria (\S\ref{sec:rubric_evaluation}).

\begin{table}[ht]
\centering
\caption{Comparison of \benchname{} with shopping-domain benchmarks
and rubric-based benchmarks in other domains. Eval-item counts appear below each name. ``MT'' = multi-turn; ``Expert'' = expert-authored; ``Rubric'' = expert-authored rubric scoring. }
\label{tab:benchmark_comparison}
\setlength{\tabcolsep}{3pt}
\small
\renewcommand{\arraystretch}{1.2}
\begin{tabular*}{\columnwidth}{@{}p{4.8cm}@{\extracolsep{\fill}}w{c}{0.5cm}w{c}{0.9cm}w{c}{0.8cm}@{}}
\toprule
\textbf{Benchmark} & \textbf{MT} & \textbf{Expert} & \textbf{Rubric} \\
\midrule
\multicolumn{4}{l}{\textbf{\emph{Shopping-domain benchmarks}}} \\[4pt]
WebShop~\citep{yao2022webshop}\newline{\footnotesize\color{gray} (12{,}087 instructions)}   & -- & -- & -- \\[2pt]
Shopping MMLU~\citep{jin2024shoppingmmlu}\newline{\footnotesize\color{gray} (57 tasks)}      & -- & -- & -- \\[2pt]
eCeLLM~\citep{peng2024ecellm}\newline{\footnotesize\color{gray} (10 tasks)}                 & -- & -- & -- \\[2pt]
EcomEval~\citep{xie2025ecomeval}\newline{\footnotesize\color{gray} (37 tasks)}              & -- & \checkmark & -- \\[2pt]
ShoppingBench~\citep{wang2025shoppingbench}\newline{\footnotesize\color{gray} (3{,}310 tasks)} & -- & -- & -- \\[2pt]
SessionIntentBench~\citep{yang2025sessionintent}\newline{\footnotesize\color{gray} (8{,}980 trajectories)} & -- & -- & -- \\[2pt]
ShoppingComp~\citep{tou2025shoppingcomp}\newline{\footnotesize\color{gray} (120 tasks)}     & -- & \checkmark & \checkmark \\[2pt]
SmartShopBench~\citep{cheng2026chatshopbuddy}\newline{\footnotesize\color{gray} (120 tasks)} & -- & -- & \checkmark \\[2pt]
\midrule
\multicolumn{4}{l}{\textbf{\emph{Expert rubric benchmarks (other domains)}}} \\[4pt]
HealthBench~\citep{arora2025healthbench}\newline{\footnotesize\color{gray} (5{,}000 conversations)} & \checkmark & \checkmark & \checkmark \\[2pt]
ProfBench~\citep{wang2025profbench}\newline{\footnotesize\color{gray} (80 tasks)}           & -- & \checkmark & \checkmark \\[2pt]
PRBench~\citep{akyurek2025prbench}\newline{\footnotesize\color{gray} (1{,}100 questions)}   & -- & \checkmark & \checkmark \\[2pt]
\midrule
\benchname{} (ours)\newline{\footnotesize\color{gray} (1{,}996 turns)}                      & \checkmark & \checkmark & \checkmark \\
\bottomrule
\end{tabular*}
\end{table}

%% file: sections/shopping_taxonomy.tex
\section{A Taxonomy of Shopping Reasoning}
\label{sec:taxonomy}

\subsection{Design rationale}
\label{sec:taxonomy:rationale}

\begin{figure}[ht]
\centering
\includegraphics[width=\columnwidth]{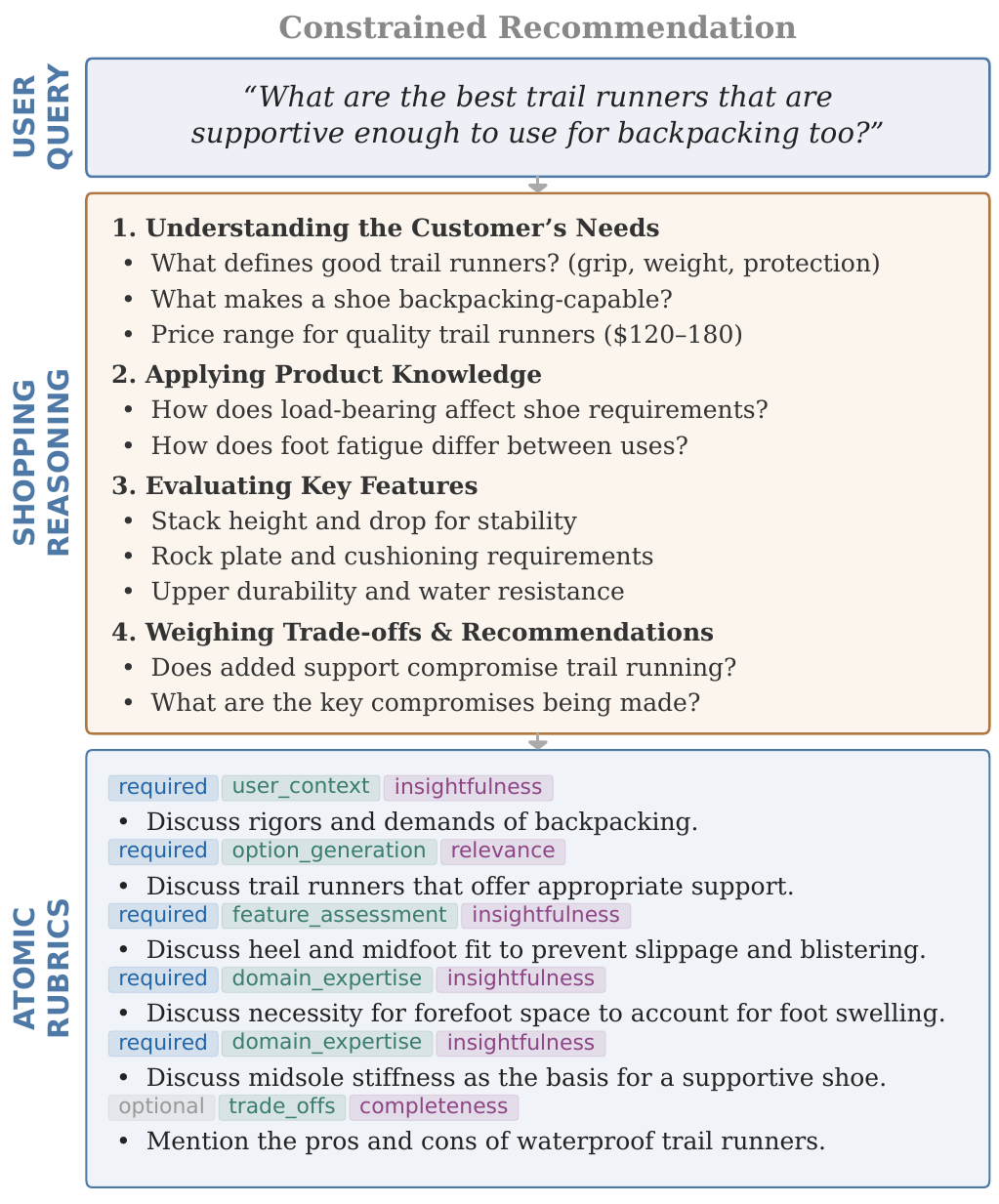}
\caption{Expert annotation pipeline illustrated on a Constrained Recommendation query from \benchname{}. The customer query is analyzed through structured reasoning stages, producing atomic rubrics---binary, independently verifiable evaluation criteria.}
\label{fig:rubric_pipeline}
\end{figure}

An expert moves through the shopping reasoning arc by understanding what a customer needs, identifying relevant options, applying domain knowledge to evaluate those options against the customer's constraints, and synthesizing actionable guidance. Figure~\ref{fig:rubric_pipeline} illustrates this arc on a representative query: an expert decomposes the query through reasoning stages and produces atomic rubrics that any adequate response must satisfy.

Existing shopping-related taxonomies target two axes: \emph{search-query intent}~\citep{sondhi2018taxonomy} and \emph{product-QA type}~\citep{yang2024bespoke}.  Both leave the actual \emph{reasoning demand} unspecified.  A query such as ``is it better to wear hiking boots or trail running shoes when thru-hiking?'' is informational in intent and comparative in form, yet the capability it probes, trade-off analysis under implicit budget and use-case constraints, is invisible to both axes. Our taxonomy targets this layer directly.

The \benchname{} taxonomy accordingly operates at two levels. At the \emph{turn level} (\S\ref{sec:taxonomy:turn}), each query is assigned to one of five reasoning categories that capture the cognitive task the query places on the assistant. At the \emph{rubric level} (\S\ref{sec:taxonomy:rubric}), every atomic rubric carries tags for reasoning stage and quality dimension that serve as analytical keys for fine-grained diagnosis. This two-level design makes the taxonomy load-bearing: rubrics constructed from reasoning-stage decomposition are reusable across product domains, and a model's capability profile across categories reveals reasoning gaps that a product-domain split would mask.

\subsection{Reasoning categories}
\label{sec:taxonomy:turn}

We defined five top-level categories to capture the dominant reasoning patterns observed in conversational shopping, and refined each into three fine-grained subcategories by requiring every leaf to support a distinct rubric template instantiated against the shopping mission context (Figure~\ref{fig:taxonomy_distributions}; definitions in Appendix~\ref{app:taxonomy}). Retail domain experts verified the mapping of each of the 1{,}996 queries and turns to a taxonomy leaf.

Two categories cover roughly 70\% of turns. \textbf{Product Recommendation} (42.8\%) ranges from narrowly constrained requests through multi-product curation to open-ended discovery, loading heavily on option generation and feature assessment. \textbf{Shopping Guidance} (26.6\%) captures queries seeking advice or education rather than product suggestions, loading on domain expertise and actionability. Three smaller categories capture distinct reasoning patterns: \textbf{Product Comparison} (10.7\%) demands weighing trade-offs across alternatives; \textbf{Product Inquiry} (10.4\%) demands depth on a single product with 54\% of its rubrics on feature assessment; and \textbf{Conversational Navigation} (9.5\%) steers the dialogue rather than requesting products---confirmed in \S\ref{sec:results} as the hardest category for every model.


\subsection{Rubric dimensions and dataset composition}
\label{sec:taxonomy:rubric}
\label{sec:taxonomy:coverage}

The benchmark comprises 1{,}996 evaluation points (232 single-turn + 1{,}764 multi-turn across 293 missions) assessed against 10{,}863 importance-weighted atomic rubrics (median~5 per turn). Each rubric carries three orthogonal tags. A \textbf{reasoning stage} identifies which step of the expert reasoning arc the rubric tests---the top three stages,  \emph{Feature Assessment} (23.3\%), \emph{Domain Expertise} (21.6\%), and \emph{Option Generation} (21.2\%), account for two-thirds of all rubrics. A \textbf{quality dimension} identifies which property of response quality is evaluated---\emph{Concreteness} (26.0\%) is the most frequently tested. \textbf{Importance} marks a rubric as \emph{required} (85\%) or \emph{optional} (15\%), separating adequacy from expert-level proactive guidance. Queries span five \textbf{product families} (\emph{hardlines} 40.6\%, \emph{softlines} 15.1\%, \emph{consumables} 14.6\%, \emph{media} 5.6\%, \emph{mixed} 24.1\%) and three \textbf{mission types} (\emph{Explore \& Discover} 57.0\%, \emph{Compare \& Choose} 22.5\%, \emph{Find Specific Solution} 20.5\%; length 2--10 turns, median~6). Full definitions appear in Appendix~\ref{app:taxonomy}.

\begin{figure}[ht]
    \centering
    \includegraphics[width=\columnwidth]{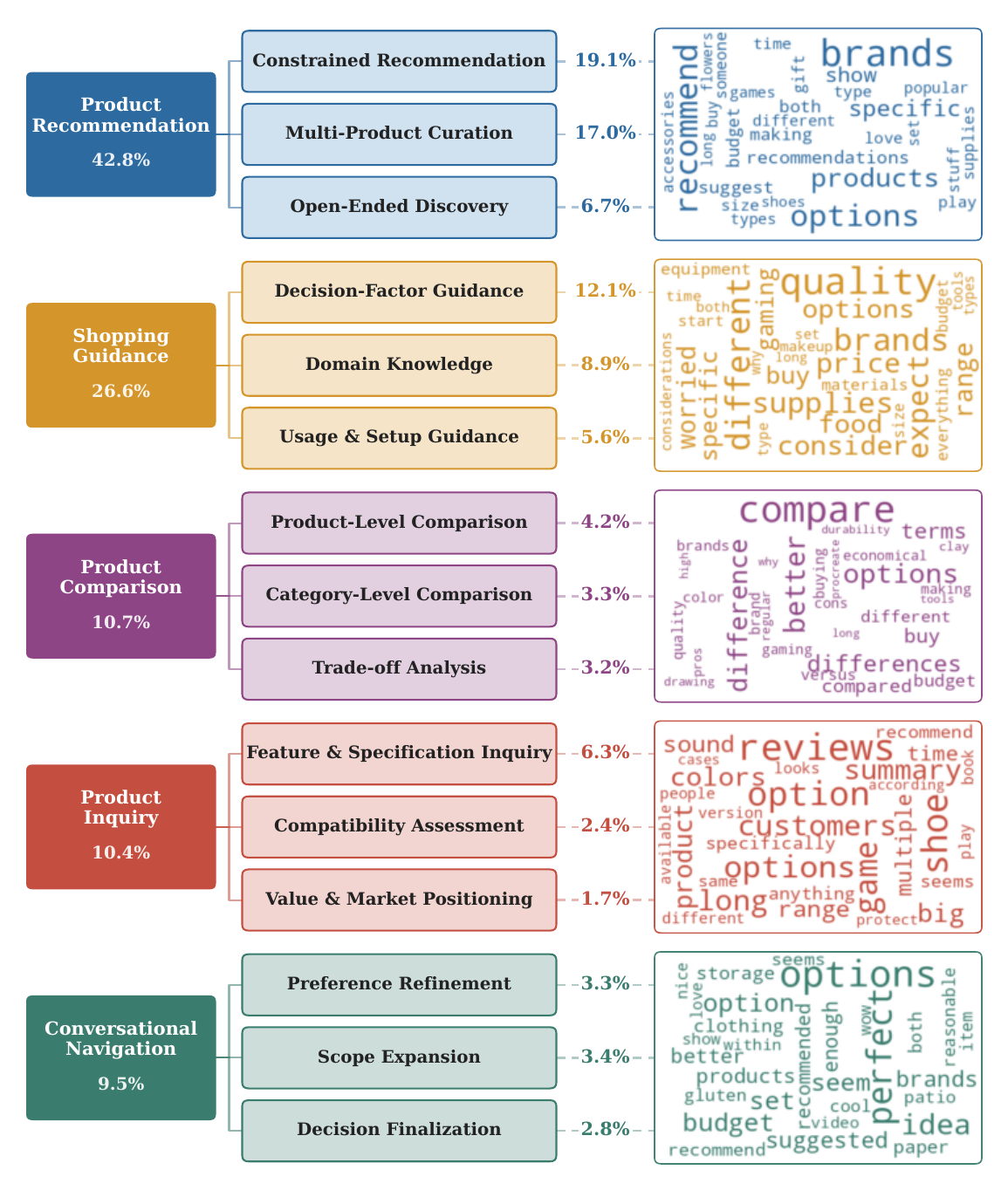}
    \caption{The \benchname{} reasoning taxonomy: five top-level categories and
fifteen fine-grained subcategories with occurrence frequencies and per-category lexical word clouds.}
    \label{fig:taxonomy_distributions}
\end{figure}

%% file: sections/rubric_evaluation.tex
\section{Evaluation Framework}
\label{sec:rubric_evaluation}
\benchname{} aggregates atomic rubric judgments
(\S\ref{sec:taxonomy:rubric}) into per-turn, per-mission, and
dataset-level scores via importance-weighted pass rates. Each rubric
is scored by a single LLM judge whose reliability is validated against
expert annotations (\S\ref{sec:eval:validation}).

\subsection{Pass rate scoring}
\label{sec:eval:scoring}

The \textbf{weighted pass rate} for a model response is
\begin{equation}
    \passrate = \frac{\sum_{i=1}^{N} w_i \cdot \mathbf{1}[\text{rubric}_i \text{ passes}]}{\sum_{i=1}^{N} w_i}
    \label{eq:pass_rate}
\end{equation}
where $N$ is the number of rubrics for a single turn, $w_i$ is the importance weight ($w_i = 5$ for required rubrics, $w_i = 1$ for optional), and $\mathbf{1}[\cdot]$ is the indicator function. Scores aggregate hierarchically: Eq.~\ref{eq:pass_rate} produces a per-turn score; the per-mission score is the arithmetic mean of its per-turn scores; the dataset-level score is the arithmetic mean of per-mission scores. This macro-average weights each turn equally within its mission and each mission equally within the dataset, so longer missions and turns with more rubrics do not dominate the aggregate.

\subsection{LLM-as-judge}
\label{sec:eval:judge}

\benchname{} uses Claude Sonnet~4.5 as the judge with fixed inference
parameters (temperature~0, single sample per rubric). A single judge
applies uniform decision criteria across the benchmark and permits
direct validation against expert annotations
(\S\ref{sec:eval:validation}).

The judge produces a binary pass/fail decision with a brief rationale
per rubric. For single-turn queries, it receives the query, model
response, and rubric text. For multi-turn evaluation, it additionally
receives the conversation history through the current turn. Prompts
and output schema are in Appendix~\ref{app:judge_prompts}; full judge and generation parameters are in \Cref{app:model_details}.

\subsection{Judge validation}
\label{sec:eval:validation}
\label{sec:judge}  

Two retail-domain experts independently labeled a stratified sample of 1{,}457 rubric instances (details in Appendix~\ref{app:annotation}). Table~\ref{tab:judge_validation} reports agreement at two levels.

\paragraph{Rubric level.}
\label{sec:eval:validation:rubric}
Each rubric is a binary \emph{met}/\emph{not-met} judgment. We report macro-F1 (mean of per-class F1, insensitive to class imbalance) and Cohen's $\kappa$. Overall macro-F1 is 0.749 ($\kappa = 0.498$, moderate; \citealp{landis1977}). The judge approaches the inter-expert ceiling---agreement between the two human experts on the same sample---for Product Recommendation and Conversational Navigation, where ceiling F1 is itself low ($\leq 0.764$). The largest gap appears in Product Comparison (0.721 vs.\ ceiling 0.852), suggesting comparative rubrics admit greater annotator subjectivity.

\paragraph{Aggregate level.}
\label{sec:eval:validation:aggregate}
We correlate the judge's importance-weighted pass-rates with experts' holistic 1--5 Likert ratings (collected per turn and per shopping mission) via Spearman's $\rho$; the inter-expert baseline replaces the judge's scores with the second expert's pass-rates. Response-level $\rho = 0.444$ ($n = 305$; baseline 0.398); mission-level $\rho = 0.469$ ($n = 30$; baseline 0.389). The judge slightly exceeds the inter-expert baseline at both levels.

\begin{table*}[t]
\centering
\caption{Judge validation against expert annotations by reasoning category. Left: rubric-level macro-F1 and Cohen's $\kappa$ (judge vs.\ mission-owner expert; ceiling = inter-expert). Right: Spearman $\rho$ between judge
weighted pass-rates and expert Likert (baseline = second expert's pass-rates). $N$ = rubrics; $n$ = responses.}
\label{tab:judge_validation}
\small
\setlength{\tabcolsep}{4pt}
\begin{tabular}{@{}lrrrr@{\hspace{12pt}}rcc@{}}
\toprule
 & \multicolumn{4}{c@{\hspace{12pt}}}{\textbf{Binary agreement}} & \multicolumn{3}{c}{\textbf{Rank correlation}} \\
\cmidrule(lr){2-5} \cmidrule(l){6-8}
\textbf{Category} & $N$ & \textbf{Judge $F_1$} & \textbf{Judge $\kappa$} & \textbf{Ceiling F1\,/\,$\kappa$} & \textbf{$n$}& \textbf{Judge $\rho$} & \textbf{Inter-expert $\rho$} \\
\midrule
Product Recommendation & 637 & 0.761 & 0.523 & 0.760\,/\,0.521 & 139 & 0.381 & 0.247 \\
Shopping Guidance       & 251 & 0.731 & 0.464 & 0.766\,/\,0.534 &  57 & 0.345 & 0.313 \\
Product Comparison     & 225 & 0.721 & 0.442 & 0.852\,/\,0.704 &  44 & 0.362 & 0.578 \\
Product Inquiry        & 130 & 0.751 & 0.503 & 0.883\,/\,0.765 &  28 & 0.618 & 0.539 \\
Conversational Navigation     & 214 & 0.738 & 0.476 & 0.764\,/\,0.528 &  37 & 0.608 & 0.451 \\
\midrule
Overall              & 1{,}457 & 0.749 & 0.498 & 0.787\,/\,0.573 & 305 & 0.444 & 0.398 \\
\bottomrule
\end{tabular}
\end{table*}

%% file: sections/results.tex
\section{Results}
\label{sec:results}

We evaluate nine commercial LLMs spanning three model families (GPT, Claude, Gemini) and three capability tiers (frontier, mid, small) on \benchname{}. Each model generates responses using its native web search tool at default inference parameters (\Cref{app:model_details}). A single Claude Sonnet~4.5 judge scores all responses against the atomic rubrics; its reliability is validated against expert annotations (\Cref{sec:eval:validation}), and a cross-judge comparison with DeepSeek~V3.2 confirms that the reported rankings are robust to judge choice (\Cref{app:cross_judge}). All results use weighted pass rates (Eq.~\ref{eq:pass_rate}). Ablations on system prompt conditioning are in \Cref{app:sysprompt}.

\subsection{Main Results}
\label{sec:results_main}

\begin{table}[t]
\centering
\caption{Main results on \benchname{}. Weighted pass rate (\%) on single-turn (ST, 232 missions) and multi-turn (MT, 293 missions) subsets. Overall averages ST and MT weighted by mission count.}
\label{tab:main_results}
\small
\begin{tabular}{llccc}
\toprule
\textbf{Family} & \textbf{Model} & \textbf{ST} & \textbf{MT} & \textbf{Overall} \\
\midrule
\multirow{3}{*}{GPT}       & GPT-5.4              & 69.2 & 71.0 & 70.2 \\
                            & GPT-5.4 mini         & 61.6 & 65.8 & 63.9 \\
                            & GPT-5.4 nano         & 65.2 & 61.9 & 63.4 \\
\midrule
\multirow{3}{*}{Claude}    & Claude Opus 4.7      & 75.1 & 78.5 & 77.0 \\
                            & Claude Sonnet 4.5    & 65.1 & 71.3 & 68.6 \\
                            & Claude Haiku 4.5     & 55.3 & 59.1 & 57.4 \\
\midrule
\multirow{3}{*}{Gemini}    & Gemini 3.1 Pro       & 76.5 & 77.7 & 77.2 \\
                            & Gemini 3 Flash       & 75.2 & 75.7 & 75.5 \\
                            & Gemini 3.1 Flash-Lite & 71.1 & 73.5 & 72.4 \\
\bottomrule
\end{tabular}
\end{table}

Three properties of the benchmark emerge from the primary evaluation (\Cref{tab:main_results}). First, \benchname{} is unsaturated: overall pass rates range from 57.4\% to 77.2\%, and no model exceeds 79\% on either split. Second, the benchmark separates capability tiers: within every family, frontier models outperform mid-tier models, which in turn outperform small-tier models. Third, the two frontier models---Claude Opus~4.7 (77.0\%) and Gemini~3.1~Pro (77.2\%)---achieve comparable performance at the top of the range, while the GPT family trails at the frontier tier (70.2\%), leaving substantial room for improvement remains across all families.




\subsection{Where Do Models Struggle?}
\label{sec:results_failures}

\begin{table}[t]
\centering
\caption{Multi-turn weighted pass rate (\%) by taxonomy dimension for the frontier model of each family, averaged across turns within each group.}
\label{tab:taxonomy_breakdown}
\small
\setlength{\tabcolsep}{3pt}
\begin{tabular}{@{}lccc@{}}
\toprule
\textbf{Dimension} & \makecell{\textbf{GPT}\\\textbf{5.4}} & \makecell{\textbf{Opus}\\\textbf{4.7}} & \makecell{\textbf{Gemini}\\\textbf{3.1 Pro}} \\
\midrule
\multicolumn{4}{@{}l}{\textit{By reasoning category}} \\
Product Recommendation       & 69.3 & 76.2 & 76.9 \\
Shopping Guidance             & 74.4 & 81.5 & 79.2 \\
Product Comparison            & 72.8 & 80.9 & 77.7 \\
Product Inquiry               & 69.1 & 79.1 & 76.7 \\
Conversational Navigation     & 65.2 & 73.8 & 75.5 \\
\midrule
\multicolumn{4}{@{}l}{\textit{By product family}} \\
Hardlines                     & 68.6 & 78.4 & 76.7 \\
Softlines                     & 72.9 & 76.6 & 78.7 \\
Consumables                   & 71.8 & 80.3 & 78.1 \\
Media                         & 76.5 & 75.4 & 81.0 \\
Mixed                         & 70.4 & 78.0 & 76.9 \\
\midrule
\multicolumn{4}{@{}l}{\textit{By mission type}} \\
Explore \& Discover           & 72.4 & 78.8 & 79.1 \\
Compare \& Choose             & 67.2 & 78.0 & 74.8 \\
Find Specific Solution        & 69.1 & 76.8 & 75.6 \\
\bottomrule
\end{tabular}
\end{table}


Reasoning category produces a consistent difficulty ordering across all nine models (Table~\ref{tab:taxonomy_breakdown}). Shopping Guidance sits at the easy end: its queries lean toward advisory or educational responses, which models handle reliably. Conversational Navigation sits at the hard end: its turns mark shifts in the shopping journey, such as refining preferences as new products surface or narrowing toward a final decision, where the assistant has to re-anchor its recommendation to the customer's evolving intent. Product family shows no consistent difficulty ordering across models: each model has its own strongest and weakest product families, and no product family is uniformly hard for all nine models. Mission type yields small within-model gaps, suggesting difficulty comes from individual turns rather than mission shape.


\subsection{Required vs.\ Optional Criteria}
\label{sec:results_required_optional}

\benchname{}'s rubrics separate \emph{required} criteria (baseline shopping correctness) from \emph{optional} criteria (expert-flagged above-and-beyond advice, \S\ref{sec:taxonomy:rubric}). Every model scores 13 to 29 points lower on optional rubrics than on required ones (\Cref{tab:req_opt}). Current models cover the basics of shopping assistance at a reasonable rate but less consistently produce the kind of above-and-beyond advice that domain experts consider the mark of high-quality assistance.

\begin{table}[t]
\centering
\caption{Multi-turn required vs.\ optional rubric pass rates (\%), averaging the per-turn fraction of rubrics met within each importance class. Gap = Optional $-$ Required; negative values indicate that models perform worse on above-and-beyond criteria.}
\label{tab:req_opt}
\small
\begin{tabular}{llccc}
\toprule
\textbf{Family} & \textbf{Model} & \textbf{Req.} & \textbf{Opt.} & \textbf{Gap} \\
\midrule
\multirow{3}{*}{GPT}       & GPT-5.4              & 71.6 & 46.5 & $-$25.1 \\
                            & GPT-5.4 mini         & 66.8 & 37.8 & $-$29.0 \\
                            & GPT-5.4 nano         & 62.6 & 37.2 & $-$25.4 \\
\midrule
\multirow{3}{*}{Claude}    & Claude Opus 4.7      & 78.8 & 66.0 & $-$12.8 \\
                            & Claude Sonnet 4.5    & 72.0 & 50.8 & $-$21.2 \\
                            & Claude Haiku 4.5     & 60.2 & 36.9 & $-$23.3 \\
\midrule
\multirow{3}{*}{Gemini}    & Gemini 3.1 Pro       & 78.3 & 58.0 & $-$20.3 \\
                            & Gemini 3 Flash       & 76.0 & 58.7 & $-$17.3 \\
                            & Gemini 3.1 Flash-Lite & 73.9 & 56.8 & $-$17.1 \\
\bottomrule
\end{tabular}
\end{table}

\subsection{Rubric Difficulty Distribution by Reasoning Dimension}
\label{sec:results_difficulty}

The previous sections show \emph{where} models struggle by category and by importance class. Table~\ref{tab:rubric_difficulty} provides a finer-grained view, breaking down rubric difficulty by reasoning stage and quality dimension (definitions in \Cref{app:taxonomy}). Of the 10{,}042 multi-turn rubrics, 28.3\% are passed by all nine models (ceiling), 5.3\% are passed by none (floor), and the remaining 66.4\% discriminate between models.

\begin{table}[t]
\centering
\caption{Rubric difficulty by dimension. Floor = fraction passed by no model; Ceiling = fraction passed by all nine models. Higher ceiling indicates easier rubrics; higher floor indicates harder rubrics.}
\label{tab:rubric_difficulty}
\small
\begin{tabular}{lrrr}
\toprule
\textbf{Dimension} & \textbf{$N$} & \textbf{Floor (\%)} & \textbf{Ceiling (\%)} \\
\midrule
\multicolumn{4}{l}{\textit{By importance}} \\
Required            & 8{,}502 & 4.0  & 31.3 \\
Optional            & 1{,}540 & 12.3 & 11.6 \\
\midrule
\multicolumn{4}{l}{\textit{By reasoning stage}} \\
User Context        &   699 & 3.1 & 47.1 \\
Trade-offs          &   806 & 4.8 & 32.9 \\
Option Generation   & 2{,}143 & 5.6 & 28.9 \\
Domain Expertise    & 2{,}063 & 4.8 & 26.7 \\
Feature Assessment  & 2{,}303 & 4.1 & 21.8 \\
Actionability       & 2{,}028 & 7.8 & 28.4 \\
\midrule
\multicolumn{4}{l}{\textit{By reasoning quality}} \\
Clarity             &   744 & 6.3 & 49.5 \\
Relevance           & 2{,}192 & 4.0 & 35.8 \\
Accuracy            &   907 & 3.6 & 28.6 \\
Completeness        & 1{,}905 & 5.8 & 29.0 \\
Concreteness        & 2{,}754 & 5.9 & 21.4 \\
Insightfulness      & 1{,}540 & 5.9 & 18.8 \\
\bottomrule
\end{tabular}
\end{table}

The importance split reinforces the required--optional gap from \S\ref{sec:results_required_optional}: 31.3\% of required rubrics are passed by all nine models, compared to only 11.6\% of optional rubrics. 

Among reasoning stages, \emph{user context} rubrics are the easiest (47.1\% passed by all models): models reliably identify what the customer is asking for. \emph{Feature assessment} has the lowest all-pass rate (21.8\%), indicating that evaluating specific product attributes (materials, compatibility, specifications) is where models diverge most. \emph{Actionability} rubrics show a distinct pattern: despite a moderate all-pass rate (28.4\%), they have the highest rate of rubrics that no model passes (7.8\%), suggesting that giving concrete, usable recommendations is an area of consistent difficulty.

Among quality dimensions, \emph{clarity} (49.5\% all-pass) and \emph{relevance} (35.8\%) are well-handled, while \emph{insightfulness} (18.8\%) and \emph{concreteness} (21.4\%) show the lowest all-pass rates. Models produce organized, on-topic responses but less consistently demonstrate expert-level depth or provide specific, tangible details. In the shopping domain, this gap matters: the difference between a generic answer and a useful one often lies in the concrete product knowledge that requires domain expertise to produce.

\subsection{Multi-Turn Degradation}
\label{sec:results_multiturn}

All three frontier models show declining pass rates as missions progress (\Cref{app:turn_degradation}). GPT-5.4 drops 10.3 points (77.4 → 67.1), Gemini 3.1 Pro drops 7.3 points (81.5 → 74.2), and Claude Opus 4.7 drops least at 4.5 points (78.5 → 74.0). Three frontier models degrading at different rates on the same missions confirms that sustained multi-turn coherence is a distinct capability, not a function of single-turn quality.

%% file: sections/discussion.tex
\section{Discussion}
\label{sec:discussion}

\paragraph{The taxonomy as an evaluation lens.}
Organizing evaluation by reasoning category reveals capability gaps that simpler breakdowns obscure. Conversational Navigation is consistently the hardest category across all three families, while Shopping Guidance is consistently the easiest. This pattern holds across all three families despite large differences in overall score. A product-family breakdown, by contrast, produces model-specific patterns with no consistent ordering (Table~\ref{tab:taxonomy_breakdown}). The taxonomy-based decomposition is more diagnostic because it groups turns by the cognitive demand they place on the assistant (e.g., trade-off reasoning, product knowledge retrieval) rather than by surface-level topic. Rubric-level tags extend this further: breaking down pass rates by reasoning stage and quality dimension (\Cref{app:st_mt_stage}) shows that models handle user context recognition well but struggle with actionability and insightfulness, a distinction invisible in category-level or product-level aggregates.

\paragraph{What importance weighting reveals.}
The 13--29 point gap between required and optional rubric pass rates (\Cref{sec:results_required_optional}) is \benchname{}'s most distinctive empirical finding. Required rubrics test whether a response covers what the customer asked for; optional rubrics test whether it goes further with proactive advice, complementary suggestions, or decision frameworks. Without importance weighting, these two classes would be averaged together, compressing the quality spectrum and making adequate responses look closer to expert-level ones than they are. The gap shows that current models reliably cover the basics of shopping assistance but less consistently produce the above-and-beyond advice that domain experts consider the mark of high quality. This has a design implication for rubric benchmarks beyond shopping: distinguishing must-have from nice-to-have criteria, and weighting them differently, exposes a capability dimension that binary pass/fail alone would miss. A within-stage breakdown of this gap is in \Cref{app:gap_by_stage}.

%% file: sections/conclusion.tex
\section{Conclusion}
\label{sec:conclusion}

We introduced \benchname{}, an expert-authored benchmark for evaluating multi-turn shopping assistance. Retail shopping poses distinctive evaluation challenges: subjective preference resolution, cross-product trade-off reasoning, and multi-turn purchase-decision progression. These require dedicated expert-crafted criteria rather than general-purpose metrics. \benchname{} addresses this with 525 missions across five product families, scored against 10{,}863 importance-weighted atomic rubric criteria organized around the first taxonomy of pre-purchase shopping reasoning (5 categories, 15 subcategories).

Our evaluation of nine models from three families shows that the benchmark is both unsaturated and discriminative. Pass rates range from 57\% to 77\%, and all models score 13--29 points lower on optional rubrics than on required ones. The hardest shopping-specific skills remain far from solved. The rubric decomposition reveals that models handle the basics but fall short on the above-and-beyond criteria that domain experts consider the mark of high-quality advice. By grounding evaluation in expert-authored atomic rubrics with importance weights, \benchname{} provides the resolution needed to measure capability gains from domain-specific post-training of shopping assistants. We release the full benchmark together with a focused \benchnameHard{} subset of the 108 hardest missions (Appendix~\ref{app:benchmark_variants}).

%% file: sections/limitations.tex
\section*{Limitations}

On the evaluation side, each model generated one response per query (no repeated sampling), so scores reflect a single draw from each model's output distribution. Variance across samples could affect individual mission scores, though dataset-level averages over 525 missions mitigate this. Some per-category and per-family breakdowns are based on small subsets (e.g., Media with 14 multi-turn missions) and should be interpreted cautiously.

On the data side, the expert team covers five product families but several domains (grocery, automotive, industrial supplies, digital services) are not represented. Some taxonomy labels (e.g., query type, shopping funnel stage) were calibrated through iterative expert review rather than single-pass annotation. While all labels were verified for accuracy, systematic biases from the calibration process may propagate into the taxonomy distribution. Product knowledge evolves over time: new products launch, prices change, and availability fluctuates. The 17.7\% of single-turn queries and 38.9\% of multi-turn turns flagged as time-sensitive may become outdated, requiring periodic benchmark updates.

%% file: sections/ethics.tex
\section*{Ethics Statement}

\paragraph{Data collection.}
All benchmark queries were authored by expert annotators who were informed of the research purpose. No customer data, personally identifiable information, or proprietary product data was used.

\paragraph{Potential misuse.}
While \benchname{} is designed to evaluate and improve shopping assistants, the rubric framework could in principle be used to optimize models for persuasive or manipulative product recommendations. We encourage responsible use focused on improving response accuracy and helpfulness rather than maximizing conversion.

\paragraph{Bias considerations.}
The benchmark reflects the knowledge and perspectives of 7 domain experts and may inherit biases related to product preferences, brand familiarity, and cultural context. We encourage users to consider these limitations when interpreting evaluation results.

\paragraph{Environmental impact.}
LLM-based evaluation requires significant computational resources. The single-turn subset (232 queries, 821 rubrics) can serve as a lightweight evaluation where full multi-turn assessment is not required.

\paragraph{Use of AI assistants.}
AI writing assistants were used for editorial refinement of the manuscript. All content has been audited and modified by the authors.

\section*{Data Availability}
The benchmark data (queries, missions, rubrics, and taxonomy), annotation guidelines, and judge prompts will be released under the Creative Commons Attribution-NonCommercial 4.0 International License (CC-BY-NC-4.0) upon acceptance. An access-gated release ensures responsible use while maintaining reproducibility.

%% file: appendices/appendix_taxonomy.tex
\section{Taxonomy and Rubric Definitions}
\label{app:taxonomy}

This appendix provides complete definitions for all taxonomy dimensions and rubric design principles used in \benchname{}.

\subsection{Product Families}

\begin{description}[style=nextline,leftmargin=2em]
    \item[Hardlines] Durable goods including electronics, appliances, tools, sports equipment, furniture, and home improvement.
    \item[Softlines] Apparel, footwear, accessories, textiles, and fashion items.
    \item[Consumables] Food, beverages, health and beauty products, household supplies, and other consumable goods.
    \item[Media] Books, music, movies, video games, and digital content.
    \item[Mixed] Queries spanning multiple product families.
\end{description}

\subsection{Mission Types (Multi-Turn Only)}

\begin{description}[style=nextline,leftmargin=2em]
    \item[Explore \& Discover] Open-ended shopping journeys where customers browse, learn about options, and gradually narrow their preferences over multiple turns.
    \item[Compare \& Choose] Focused comparison shopping between specific products or categories, leading to a selection decision.
    \item[Find Specific Solution] Goal-directed shopping for a particular need, problem, or use case with relatively clear requirements.
\end{description}

\subsection{Shopping Funnel Stages}

Each turn in a multi-turn mission is labeled with one of three funnel stages reflecting the customer's shopping intent at that point in the conversation. Percentages below are computed over all 1{,}764 multi-turn turns. At the mission level, the ordered sequence of per-turn labels is stored as the \texttt{shopping\_funnel\_flow} array.

\begin{description}[style=nextline,leftmargin=2em]
    \item[Discover (31.4\%)] Broadly exploring needs---the customer is in the early stage with undefined or loosely defined intent.
    \item[Explore (62.9\%)] Evaluating specific options---the customer has narrowed the space and is comparing or learning about particular products or categories.
    \item[Ready-to-Transact (5.7\%)] Finalizing a decision---the customer is close to or at the point of purchase.
\end{description}

\subsection{Reasoning Stage and Quality Dimensions}
\label{app:taxonomy:stage_quality}

Table~\ref{tab:stage_quality_defs} gives the full definitions of the six reasoning stages and six quality dimensions used to tag each of the 10{,}863 rubrics in \benchname{}.

\begin{table*}[ht]
\centering
\caption{Definitions of the six reasoning stages and six quality dimensions used to tag each rubric. Percentages give the share of \benchname{}'s 10{,}863 rubrics carrying each tag.}
\label{tab:stage_quality_defs}
\small
\setlength{\tabcolsep}{4pt}
\begin{tabular}{@{}llr@{}}
\toprule
\textbf{Tag} & \textbf{Definition} & \textbf{\%} \\
\midrule
\multicolumn{3}{l}{\emph{Reasoning stage}} \\
\midrule
User Context        & Interprets the customer's situation, constraints, and intent     & 7.1 \\
Option Generation   & Surfaces relevant products or categories                         & 21.2 \\
Domain Expertise    & Requires specialized product knowledge                           & 21.6 \\
Feature Assessment  & Evaluates specific attributes and specifications                 & 23.3 \\
Trade Offs          & Comparison and prioritization reasoning                          & 8.0 \\
Actionability       & Recommendations are concrete and usable                          & 18.9 \\
\midrule
\multicolumn{3}{l}{\emph{Reasoning quality}} \\
\midrule
Concreteness        & Recommendations include specific, tangible details               & 26.0 \\
Relevance           & Content addresses actual customer needs                          & 22.3 \\
Completeness        & All important aspects are covered                                & 20.0 \\
Insightfulness      & Expert-level understanding is demonstrated                       & 15.6 \\
Accuracy            & Facts and specifications are correct                             & 9.2 \\
Clarity             & Information is well-organized                                    & 6.8 \\
\bottomrule
\end{tabular}
\end{table*}

\input{sections/rubric_taxonomy}

%% file: sections/rubric_taxonomy.tex
\subsection{Rubric Taxonomy}
\label{sec:rubric_taxonomy}

The four rubric tag dimensions are defined in \S\ref{sec:taxonomy:rubric}; this subsection provides design principles, the stage--quality cross-tabulation, and per-category reasoning stage profiles.

\subsubsection{Rubric Design Principles}
\label{sec:rubric_principles}

Drawing on operational experience from expert annotation across all five reasoning categories, we identify seven principles that characterize effective evaluation rubrics:

\begin{enumerate}
    \item \textbf{Clear \& Unambiguous.} Each rubric uses precise language so that both human annotators and LLM judges reach the same pass/fail decision. Vague terms like ``good'' or ``appropriate'' are replaced with specific, measurable criteria.

    \item \textbf{Actionable.} Rubrics describe observable response behaviors rather than internal model states. A judge can determine pass/fail by examining the response text alone.

    \item \textbf{Comprehensive.} The rubric set for a query covers the reasoning stages that the query's reasoning category demands---for example, Product Comparison rubrics emphasize feature assessment and trade-off reasoning, while Shopping Guidance rubrics emphasize domain expertise and actionability.

    \item \textbf{Aligned with System Capabilities.} Rubrics do not penalize responses for limitations outside the model's control (e.g., real-time inventory checks) and account for what a text-based assistant can reasonably provide.

    \item \textbf{Balanced.} Rubrics test both the presence of required information (recall) and the absence of harmful or irrelevant content (precision), avoiding over-emphasis on either direction.

    \item \textbf{Fair.} Multiple valid response strategies can satisfy the same rubric. Rubrics avoid requiring a single ``correct'' phrasing or product ordering unless specificity is essential.

    \item \textbf{Atomic.} Each rubric tests exactly one aspect of the response. Compound criteria (e.g., ``Recommend a durable and affordable product'') are split into separate rubrics.
\end{enumerate}

In addition, we enforce the following operational guidelines:
\begin{itemize}
    \item Rubrics are ordered by importance, with required rubrics listed before optional ones.
    \item Each rubric is written as a complete sentence describing the expected response behavior.
    \item Queries contain between 2 and 10 rubrics.
    \item Rubric language targets non-expert comprehension to ensure consistent LLM judge interpretation.
\end{itemize}

\begin{table*}[t]
\centering
\caption{Cross-tabulation of reasoning stage and reasoning quality across all 10,863 rubrics. Each cell shows the number of rubrics at the intersection.}
\label{tab:stage_quality_cross}
\small
\begin{tabular}{lrrrrrr|r}
\toprule
\textbf{Stage $\backslash$ Quality} & \textbf{Accuracy} & \textbf{Complete} & \textbf{Concrete} & \textbf{Relevance} & \textbf{Insightful} & \textbf{Clarity} & \textbf{Total} \\
\midrule
User Context        &   60 &   55 &    7 &  567 &   39 &   40 &   768 \\
Option Generation   &   22 &  727 &  877 &  626 &   17 &   32 & 2,301 \\
Domain Expertise    &  584 &  526 &  274 &   40 &  803 &  117 & 2,344 \\
Feature Assessment  &  250 &  455 &  738 &  662 &  392 &   29 & 2,526 \\
Trade Offs          &   33 &  201 &  167 &  152 &  228 &   92 &   873 \\
Actionability       &   48 &  211 &  764 &  374 &  220 &  434 & 2,051 \\
\midrule
\textbf{Total}       &  997 & 2,175 & 2,827 & 2,421 & 1,699 &  744 & 10,863 \\
\bottomrule
\end{tabular}
\end{table*}

\subsubsection{Reasoning Stage Profiles by Category}
\label{sec:reasoning_category_rubrics}

Different reasoning categories produce distinct rubric distributions across the six reasoning stages. \Cref{tab:reasoning_category_rubric_summary} summarizes the stage profiles computed from all 10,863 published rubrics. Across all categories, 84--87\% of rubrics are marked required and 13--16\% optional, reflecting the benchmark's emphasis on must-have evaluation criteria.

\paragraph{Product Recommendation} (4,649 rubrics).
Covers Constrained, Multi-Product, and Open-Ended subcategories. Option generation (32\%) and feature assessment (24\%) dominate, reflecting the need to surface relevant product categories and evaluate their fit. Actionability (16\%) and domain expertise (15\%) provide supporting depth.

\paragraph{Shopping Guidance} (2,995 rubrics).
Covers Decision-Factor, Domain Knowledge, and Usage \& Setup subcategories. Domain expertise (38\%) leads by a wide margin, followed by actionability (27\%), consistent with queries that seek educational content and practical next steps rather than product lists.

\paragraph{Product Comparison} (1,195 rubrics).
Covers Product-Level, Category-Level, and Trade-off Analysis subcategories. Trade Offs (36\%) is the single largest stage---uniquely high among all categories---paired with feature assessment (22\%) and domain expertise (20\%), capturing the structured comparative reasoning these queries require.

\paragraph{Product Inquiry} (1,057 rubrics).
Covers Feature \& Spec, Compatibility, and Value \& Market subcategories. Feature assessment (54\%) dominates, the highest single-stage concentration in the benchmark, reflecting queries that target specific product attributes and specifications.

\paragraph{Conversational Navigation} (967 rubrics).
Covers Preference Refinement, Scope Expansion, and Decision Finalization subcategories. User context (15\%) is uniquely elevated for this category, while actionability (25\%), option generation (24\%), and feature assessment (24\%) ensure multi-turn coherence translates into concrete guidance.

\begin{table*}[t]
\centering
\caption{Reasoning stage distribution per category (percentage of rubrics) and importance split.}
\label{tab:reasoning_category_rubric_summary}
\small
\setlength{\tabcolsep}{4pt}
\begin{tabular}{@{}lcccccccc@{}}
\toprule
\textbf{Category} & \textbf{N} &  \makecell{\textbf{User}\\\textbf{Context}} & \makecell{\textbf{Option} \\\textbf{Generation}} & \makecell{\textbf{Domain} \\\textbf{Expertise}} & \makecell{\textbf{Feature}\\ \textbf{Assessment}} & \makecell{\textbf{Trade} \\\textbf{Offs}} & \textbf{Actionability} & \textbf{Required\%} \\
\midrule
Product Recommendation    & 4,649 &  7 & 32 & 15 & 24 &  5 & 16 & 84 \\
Shopping Guidance   & 2,995 &  5 & 14 & 38 & 11 &  5 & 27 & 84 \\
Product Comparison   & 1,195 &  4 &  7 & 20 & 22 & 36 & 11 & 87 \\
Product Inquiry   & 1,057 &  8 &  6 & 15 & 54 &  4 & 13 & 87 \\
Conversational Navigation    &   967 & 15 & 24 &  9 & 24 &  3 & 25 & 87 \\
\bottomrule
\end{tabular}
\end{table*}

%% file: appendices/appendix_annotation.tex
\section{Expert Panel, Authoring, and Annotation}
\label{app:annotation}

\subsection{Expert panel and authoring}
\label{sec:taxonomy:experts}

We recruited retail domain experts, each with product knowledge
spanning at least three categories. Selection criteria were
(i) ability to produce accurate product recommendations grounded in
technical product attributes, and (ii) familiarity with real customer
shopping patterns across multiple price points and use cases. The
panel collectively covers the five product families.

Rather than writing evaluation criteria directly, each expert follows
a structured reasoning process that decomposes ambiguous customer
needs into specific technical considerations---fit, materials,
compatibility, trade-offs---before deriving concrete rubric criteria
that any adequate response must address
(Figure~\ref{fig:rubric_pipeline}). This decomposition makes explicit
the domain knowledge that distinguishes expert shopping assistance
from keyword matching: a ``best trail runners for backpacking'' query
is not answered by retrieving popular trail-runner SKUs but by
reasoning about dual-use load-bearing requirements, midsole stiffness,
and the trade-off between trail agility and backpacking support.

\begin{figure*}[ht]
\centering
\includegraphics[width=\textwidth]{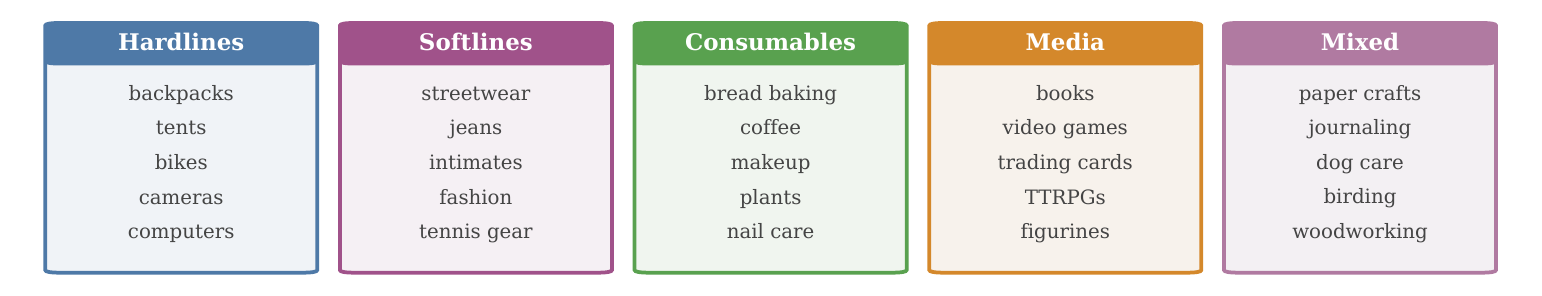}
\caption{Expert domain coverage across the five product families in
\benchname{}, with representative subtopics illustrating the breadth
of retail expertise required.}
\label{fig:domain_coverage}
\end{figure*}

\subsection{Annotation Guidelines \& Templates}
\label{app:annotation:guidelines}

This subsection describes the annotation guidelines provided to expert annotators.

\subsubsection{Single-Turn Annotation Template}

Each single-turn annotation consists of the following fields:

\begin{enumerate}
    \item \textbf{Query:} A natural language shopping question or request that a customer might ask a conversational assistant.

    \item \textbf{Rubric Dimensions:} A list of atomic, binary criteria, each with:
    \begin{itemize}
        \item \texttt{rubric\_text}: A clear, verifiable statement of what the response should include.
        \item \texttt{importance}: \texttt{required} (must be satisfied) or \texttt{optional} (bonus quality).
        \item \texttt{scope}: \texttt{instance} (specific to this query) or \texttt{cluster} (category-level).
        \item \texttt{reasoning\_stage}: The expert reasoning phase the rubric tests (e.g., \texttt{user\_context}, \texttt{option\_generation}, \texttt{actionability}, \texttt{trade\_offs}).
        \item \texttt{reasoning\_quality}: The quality dimension the rubric targets (e.g., \texttt{relevance}, \texttt{insightfulness}, \texttt{completeness}, \texttt{accuracy}).
    \end{itemize}
\end{enumerate}

\subsubsection{Multi-Turn Mission Template}

Each multi-turn mission consists of:

\begin{enumerate}
    \item \textbf{Mission Tags:} Metadata including mission ID, name, type, objective, product family, and length.

    \item \textbf{Turn Sequence:} An ordered list of customer utterances representing a realistic shopping conversation flow.

    \item \textbf{Per-Turn Annotations:} For each turn:
    \begin{itemize}
        \item Turn-level tags: \texttt{reasoning\_category}, \texttt{reasoning\_subcategory}, \texttt{shopping\_funnel\_stage}.
        \item Rubric dimensions specific to the turn's expected response, each carrying the four LLM-assigned tag dimensions: \texttt{scope}, \texttt{importance}, \texttt{reasoning\_stage}, \texttt{reasoning\_quality}.
    \end{itemize}
\end{enumerate}

\subsubsection{Rubric Writing Guidelines}

Annotators were instructed to:
\begin{itemize}
    \item Write rubrics that are \textbf{atomic}---each rubric tests exactly one aspect.
    \item Ensure rubrics are \textbf{objectively verifiable}---an LLM judge should be able to determine pass/fail.
    \item Avoid rubrics that are \textbf{too vague} (e.g., ``Response is helpful'') or \textbf{too specific} (e.g., ``Response contains exactly 5 product recommendations'').
    \item Tag rubrics as \texttt{required} if failing them would make the response fundamentally inadequate, and \texttt{optional} if they represent desirable but non-essential qualities.
    \item Aim for 2--7 rubrics per single-turn query and 3--11 per multi-turn turn.
\end{itemize}

\subsection{Judge validation protocol}
\label{app:annotation:validation}

\paragraph{Model and sample.}
Judge validation is conducted on responses produced by an evaluated model, Gemini~2.5 Pro. A two-stage stratified sampling protocol draws 124 single-turn queries and 30 multi-turn missions, yielding 1{,}457 rubric-level validation instances that span all five reasoning categories and both benchmark splits.

\paragraph{Annotator assignment and blinding.}
Each instance is labeled independently by two experts who are blinded to (i)~the identity of the model that produced the response, (ii)~the LLM judge's label, and (iii)~each other's label. The mission-owner expert's labels serve as ground truth; the second expert's labels provide an inter-expert reference.

\paragraph{Reference interpretation.}
At the rubric level, inter-expert agreement defines a \emph{ceiling}: the maximum binary agreement attainable given inherent annotator subjectivity. At the aggregate level, inter-expert correlation defines a \emph{baseline}: it quantifies how well \emph{any} rubric-based aggregation predicts the mission-owner's holistic Likert rating, since the second expert also works from rubrics rather than holistic impression.

\paragraph{Metrics.}
The primary rubric-level metric is \textbf{macro-F1 (MF1)}, defined as the unweighted average of F1 on the \emph{met} class and F1 on the \emph{not-met} class:
  \begin{equation}
  \begin{split}                                                                   \mathrm{MF1} = \tfrac{1}{2}\bigl(
    &F1_{\text{met}} + F1_{\text{not-met}}\bigr), \\
  \text{where } F1_c &= \frac{2\,TP_c}{2\,TP_c + FP_c + FN_c}.
  \end{split}
  \label{eq:mf1}
  \end{equation}
This choice ensures equal weight to both classes despite \benchname{}'s pass-heavy distribution (71.9\% \emph{met} in the validation sample). As a secondary metric we report Cohen's $\kappa$~\citep{cohen1960}, which adjusts for chance agreement. At the aggregate level, we report Spearman's $\rho$ between the judge's importance-weighted pass-rates and the mission-owner expert's 1--5 Likert ratings, computed separately at response level ($n = 305$) and mission level ($n = 30$).

\paragraph{Coverage note.}
The judge applies the full rubric set to every response, whereas each human expert annotates only their assigned subset. This asymmetry explains why the judge slightly exceeds the inter-expert baseline for rank correlation: the judge aggregates over more rubric judgments per response than any single expert.

%% file: appendices/appendix_model_details.tex
\section{Inference and Judge Parameters}
\label{app:model_details}

This appendix documents the inference parameters for evaluated models and the judge model used in \benchname{}. The nine evaluated models (three families, three capability tiers each) are listed in \Cref{tab:main_results}. For the Claude family, Opus~4.7 is used at the frontier tier while the mid and small tiers use Sonnet~4.5 and Haiku~4.5, as no 4.7-generation models are available at those tiers.

\paragraph{Generation parameters.}
All nine models---GPT-5.4 family~\citep{openai2026models}, Claude family~\citep{anthropic2026models}, and Gemini family~\citep{google2026models}---are evaluated at temperature~1.0 (the API default for all providers), with top-$p$ and maximum output tokens left at API defaults. Each model generates one response per query (single-turn) or per turn (multi-turn). All models have web search enabled via each provider's native tool integration (OpenAI web search, Google grounding, Anthropic web search). The model decides autonomously whether to invoke search on each turn; no forced-search or no-search constraint is applied.

\paragraph{Judge model.}
\benchname{} uses Claude Sonnet~4.5 as the single LLM judge at temperature~0, producing one binary pass/fail decision with a brief rationale per rubric. For single-turn queries, the judge receives the query, model response, and rubric text. For multi-turn evaluation, it additionally receives the full conversation history through the current turn. The prompt templates and output schema are documented in \Cref{app:judge_prompts}. Self-judging bias toward the Claude-family evaluated models is an inherent property of this design; a cross-judge comparison in \Cref{app:cross_judge} confirms no self-preference effect on rankings.

%% file: appendices/appendix_judge_prompts.tex
\section{LLM Judge Prompt}
\label{app:judge_prompts}

\benchname{} uses a single prompt template for both single-turn and multi-turn evaluation. For single-turn queries, the conversation history field is left blank. For multi-turn evaluation, it contains all prior turns in chronological order. Figure~\ref{fig:judge_prompt} reproduces the prompt verbatim; five worked examples are omitted for space but are included in the released evaluation code.

\begin{figure*}[t]
\lstset{
  basicstyle=\scriptsize\ttfamily,
  breaklines=true,
  breakatwhitespace=true,
  breakindent=1em,
  columns=fullflexible,
  keepspaces=true,
  showstringspaces=false,
  upquote=true,
  xleftmargin=0pt,
  xrightmargin=0pt,
  frame=single,
  framerule=0.4pt,
  rulecolor=\color{csrbgray},
  aboveskip=0pt,
  belowskip=0pt,
  postbreak=\mbox{\textcolor{csrbgray}{$\hookrightarrow$}\space},
}
\begin{lstlisting}
Your task is to evaluate the assistant's latest response in the **current conversation** between the user and the AI shopping assistant based on the given **rubric** to determine whether it meets the rubric's requirements.

During the evaluation, you may refer to the **conversation history** if the rubric requires context from earlier turns.

## Key Information to Focus On

### Current Conversation
<<current_conversation>>

### Rubric
<<rubric_text>>

### Reference Evidence
- **Conversation history**
<<conversation_history>>

---

## Instructions

### Evaluation Scope
- **PRIMARY FOCUS**: The assistant's latest response in the "Current Conversation".
- **EVALUATION BASIS**: Determine whether the assistant's latest response satisfies ALL criteria specified in the "Rubric".
- **SCOPE LIMITATION**: Only evaluate what is explicitly required by the rubric, do NOT add or consider additional criteria.

### When to Use Reference Evidence
- **General principle**:
  - Only reference the conversation history when the rubric explicitly requires verification against earlier turns.
  - If the rubric can be evaluated solely from the current conversation, consulting conversation history is optional.
- **If the conversation history is blank**:
  - This indicates the rubric can be evaluated purely from the current conversation.
  - Evaluate based solely on the assistant's response content and structure.

### Return a json object with the following fields: "explanation" and "rubric_met"
- The "explanation" field should be a string explaining why the response does or does not meet the rubric.
- The "rubric_met" field should be a boolean indicating whether the response meets the rubric.
- If rubric has multiple requirements connected by "and": ALL must be met for true. If rubric has requirements connected by "or": ANY ONE met results in true. If rubric asks for "at least X": meeting or exceeding X results in true.
- One important exception to the above bullet point is that if a rubric says "such as", "for example", "including", or "e.g.", the rubric does not have to include all of the examples listed to meet the rubric.
- For rubrics about avoiding behaviors: the response should be classified as true if it successfully avoids the undesirable behavior, and false if it exhibits the undesirable behavior.

### Edge Cases
- If the assistant's response is empty or error message: rubric_met = false.

---

# Examples
Five worked examples omitted for space. Examples cover: 
(1) cross-turn duplication detection
(2) missing concrete details
(3) context-awareness failure
(4) incorrect product identification from history
(5) successful workflow explanation.

# Output Format
- Return just the JSON object in markdown format. Do not include any other text in the response.
- "explanation": String, 1-3 sentences focusing on WHY the rubric is/isn't met
- "rubric_met": Boolean (true/false only)
\end{lstlisting}
\caption{Full LLM judge prompt used by \benchname{}. The \texttt{<<\ldots>>} tokens are filled at evaluation time with the current turn, rubric text, and (for multi-turn) conversation history through the preceding turn.}
\label{fig:judge_prompt}
\end{figure*}

%% file: appendices/appendix_results.tex
\section{Extended Results}
\label{app:extended_results}
\label{app:results}

This appendix extends the main evaluation (\S\ref{sec:results}) with a within-stage breakdown of the required--optional gap, per-family breakdowns by reasoning stage and quality, and multi-turn degradation for all nine models. Per-category and per-mission-type breakdowns for the three frontier models are reported in \Cref{tab:taxonomy_breakdown}.

\subsection{Required--Optional Gap by Reasoning Stage}
\label{app:gap_by_stage}

A natural concern is whether optional rubrics are simply harder because they test more demanding reasoning stages. Table~\ref{tab:gap_by_stage} breaks down the required--optional gap by reasoning stage, pooling rubric judgments from all nine standard models on multi-turn missions. The gap persists within every stage, ranging from $-$17.2 points (Feature Assessment) to $-$22.5 points (Domain Expertise), indicating that the gap is not an artifact of stage difficulty. The widest gaps appear in Domain Expertise ($-$22.5 points) and Actionability ($-$20.7 points), while Feature Assessment shows the narrowest ($-$17.2 points).

\begin{table}[tb]
\centering
\small
\caption{Required vs.\ optional pass rates by reasoning stage (MT, pooled across 9 models).}
\label{tab:gap_by_stage}
\begin{tabular}{lrr}
\toprule
\textbf{Stage} & \textbf{Required \%} & \textbf{Optional \%} \\
\midrule
User Context        & 79.1 & 60.8  \\
Trade Offs          & 72.4 & 55.0  \\
Option Generation   & 69.9 & 51.2  \\
Domain Expertise    & 70.3 & 47.8  \\
Feature Assessment  & 69.8 & 52.6  \\
Actionability       & 67.2 & 46.5  \\
\midrule
Overall             & 70.4 & 49.7  \\
\bottomrule
\end{tabular}
\end{table}

\subsection{Pass Rates by Reasoning Stage and Quality}
\label{app:st_mt_stage}

Table~\ref{tab:st_mt_breakdown_frontier} reports weighted pass rates by reasoning stage and quality for the frontier model of each family, separately for single-turn (ST) and multi-turn (MT) missions. Reasoning stage and quality are rubric-level tags, so each rubric within a turn may carry a different tag. Among reasoning stages, Actionability scores tend to be lower in MT than ST across all families (e.g., GPT-5.4: 82.6 ST vs.\ 61.8 MT; Claude Opus~4.7: 87.0 ST vs.\ 68.7 MT). Among quality dimensions, Insightfulness is the lowest-scoring dimension in MT for all three frontier models.

\begin{table}[t]
\centering
\caption{ST vs.\ MT weighted pass rate (\%) by reasoning stage and quality for the frontier models}
\label{tab:st_mt_breakdown_frontier}
\small
\setlength{\tabcolsep}{3pt}
\begin{tabular}{@{}lcccccc@{}}
\toprule
 & \multicolumn{2}{c}{\textbf{GPT-5.4}} & \multicolumn{2}{c}{\textbf{Opus 4.7}} & \multicolumn{2}{c}{\textbf{Gem.\ 3.1 Pro}} \\
\cmidrule(lr){2-3}
\cmidrule(lr){4-5}
\cmidrule(lr){6-7}
\textbf{Dimension} & ST & MT & ST & MT & ST & MT \\
\midrule
\multicolumn{7}{l}{\textit{By reasoning stage}} \\
User Context                   & 75.4 & 79.0 & 73.9 & 81.8 & 79.7 & 82.5 \\
Option Generation              & 77.8 & 68.3 & 82.9 & 78.1 & 80.4 & 73.1 \\
Domain Expertise               & 60.9 & 63.8 & 71.9 & 77.8 & 73.3 & 76.6 \\
Feature Assessment             & 59.6 & 67.4 & 63.2 & 77.1 & 70.4 & 78.7 \\
Trade Offs                     & 58.2 & 74.7 & 68.7 & 79.4 & 58.2 & 73.1 \\
Actionability                  & 82.6 & 61.8 & 87.0 & 68.7 & 73.9 & 66.2 \\
\midrule
\multicolumn{7}{l}{\textit{By reasoning quality}} \\
Accuracy                       & 61.1 & 69.5 & 66.7 & 78.9 & 75.6 & 75.5 \\
Clarity                        & ---  & 77.4 & ---  & 75.9 & ---  & 78.9 \\
Completeness                   & 71.9 & 67.6 & 75.6 & 82.4 & 75.2 & 70.9 \\
Concreteness                   & 60.3 & 61.6 & 82.2 & 78.8 & 75.3 & 75.5 \\
Insightfulness                 & 49.7 & 60.1 & 64.8 & 72.6 & 66.0 & 73.4 \\
Relevance                      & 72.1 & 74.2 & 71.6 & 69.3 & 74.2 & 74.6 \\
\bottomrule
\end{tabular}
\end{table}

\subsection{Multi-Turn Performance Degradation}
\label{app:turn_degradation}

\begin{figure}[H]
    \centering
    \includegraphics[width=\columnwidth]{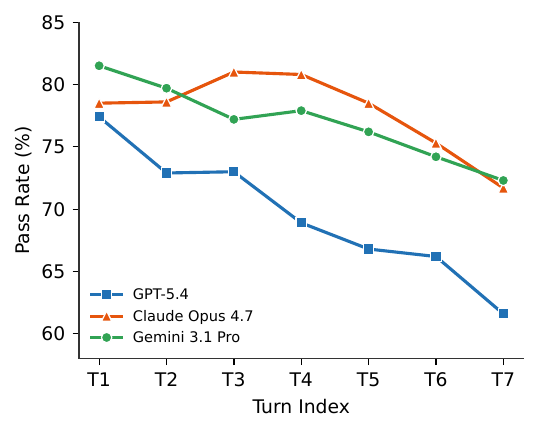}
    \caption{Per-turn pass rate for the three frontier models across turns T1 through T7.}
    \label{fig:turn_degradation}
\end{figure}

\begin{table}[t]
\centering
\caption{First-turn vs.\ last-turn weighted pass rate (\%) on multi-turn missions. For each of the 293 missions the first-turn and last-turn weighted pass rates are computed; the drop is averaged across all missions.}
\label{tab:turn_degradation}
\small
\setlength{\tabcolsep}{3pt}
\begin{tabular}{llcc}
\toprule
\textbf{Family} & \textbf{Model} & \textbf{First-turn} & \textbf{Last-turn}  \\
\midrule
\multirow{3}{*}{GPT}       & GPT-5.4              & 77.4 & 67.1 \\
                            & GPT-5.4 mini         & 71.3 & 63.6 \\
                            & GPT-5.4 nano         & 72.0 & 53.8 \\
\midrule
\multirow{3}{*}{Claude}    & Claude Opus 4.7      & 78.5 & 74.0 \\
                            & Claude Sonnet 4.5    & 73.2 & 68.6  \\
                            & Claude Haiku 4.5     & 63.0 & 57.0 \\
\midrule
\multirow{3}{*}{Gemini}    & Gemini 3.1 Pro       & 81.5 & 74.2  \\
                            & Gemini 3 Flash       & 80.2 & 70.3 \\
                            & Gemini 3.1 Flash-Lite & 76.9 & 70.2 \\
\bottomrule
\end{tabular}
\end{table}


Every model scores lower on the final turn of a mission than on the first turn (\Cref{fig:turn_degradation}, \Cref{tab:turn_degradation}). The Claude family degrades least (4.5--6.0 points), while GPT-5.4 nano is an outlier whose last-turn quality collapses by over 18 points. Within the GPT family, the frontier model degrades more than the mid-tier model, though this pattern does not hold across all families.

\subsection{Cross-Judge Validation}
\label{app:cross_judge}

To assess whether self-preference bias affects the reported rankings, we re-evaluate the three frontier models with DeepSeek~V3.2~\citep{deepseek2025v3} as an alternative judge. DeepSeek is unrelated to any of the three evaluated model families, eliminating potential same-family bias in either direction.

\begin{table}[t]
\centering
\caption{Cross-judge comparison on the three frontier models. Weighted pass rate (\%) under two judges. Rankings are preserved across judges.}
\label{tab:cross_judge}
\small
\begin{tabular}{llccc}
\toprule
\textbf{Judge} & \textbf{Model} & \textbf{ST} & \textbf{MT} & \textbf{Overall} \\
\midrule
\multirow{3}{*}{\makecell[l]{Claude\\Sonnet 4.5}}
    & GPT-5.4           & 69.2 & 71.0 & 70.2 \\
    & Claude Opus 4.7   & 75.1 & 78.5 & 77.0 \\
    & Gemini 3.1 Pro    & 76.5 & 77.7 & 77.2 \\
\midrule
\multirow{3}{*}{\makecell[l]{DeepSeek\\V3.2}}
    & GPT-5.4           & 74.8 & 82.1 & 78.9 \\
    & Claude Opus 4.7   & 80.3 & 89.3 & 85.4 \\
    & Gemini 3.1 Pro    & 81.9 & 87.9 & 85.3 \\
\bottomrule
\end{tabular}
\end{table}

Table~\ref{tab:cross_judge} shows that the two judges produce the same relative ordering: Claude Opus~4.7 and Gemini~3.1~Pro achieve comparable performance at the top, while GPT-5.4 trails both. DeepSeek is a more lenient grader overall (scores 7--9 points higher), but this shift is uniform across all three evaluated models and does not alter the ranking. These results indicate no self-preference bias in the reported evaluation.

\subsection{System Prompt Ablation}
\label{app:sysprompt}

The main evaluation uses each model's default behavior without a task-specific system prompt. To explore whether a shopping-domain system prompt affects performance, we test the three frontier models with a system prompt that defines general conditions for behaving as a shopping assistant. Results are mixed (\Cref{tab:sysprompt}): GPT-5.4 and Claude Opus~4.7 improve (+2.2 and +2.4 points overall, respectively), while Gemini~3.1~Pro decreases by 1.8 points. The divergent response suggests that base model behaviors differ in ways that interact with prompt conditioning, and that model-specific system prompt development may be needed to achieve optimal shopping assistant performance.

\begin{table}[ht]
\centering
\caption{System prompt ablation for the three frontier models. $\Delta$ is the difference relative to the default (no system prompt) condition.}
\label{tab:sysprompt}
\small
\begin{tabular}{lccc}
\toprule
\textbf{Model} & \textbf{Default} & \textbf{Sysprompt} & \textbf{$\Delta$} \\
\midrule
GPT-5.4           & 70.2 & 72.4 & $+$2.2 \\
Claude Opus 4.7   & 77.0 & 79.4 & $+$2.4 \\
Gemini 3.1 Pro    & 77.2 & 75.4 & $-$1.8 \\
\bottomrule
\end{tabular}
\end{table}

\begin{figure*}[ht]
\lstset{
  basicstyle=\scriptsize\ttfamily,
  breaklines=true,
  breakatwhitespace=true,
  breakindent=1em,
  columns=fullflexible,
  keepspaces=true,
  showstringspaces=false,
  upquote=true,
  xleftmargin=0pt,
  xrightmargin=0pt,
  frame=single,
  framerule=0.4pt,
  rulecolor=\color{csrbgray},
  aboveskip=0pt,
  belowskip=0pt,
  postbreak=\mbox{\textcolor{csrbgray}{$\hookrightarrow$}\space},
}
\begin{lstlisting}
You are an expert shopping assistant. Your role is to help customers find the right products by combining deep product knowledge with careful attention to their specific situation.

How to reason about queries:

Before responding, identify what the customer actually needs -- not just what they asked for. Infer constraints from context: their use case, environment, experience level, timeline, and budget signals. Work with what you have and give your best recommendation based on the information available.

How to respond:

- Be concrete. Name specific products, brands, models, and relevant specs. Avoid generic advice that could apply to any product.
- Commit to recommendations. Use your expertise to make judgment calls rather than deferring decisions back to the customer.
- Explain the why behind recommendations. Connect product features to the customer's actual use case and constraints.
- When multiple options exist, present them with clear trade-offs so the customer can make an informed choice.
- Match response depth to query scope. A narrow question gets a focused answer. A broad discovery question gets structured categories with examples.
- When domain knowledge is relevant (how a technology works, what makes a material durable, why a spec matters), weave it in naturally to help the customer understand their options.

In multi-turn conversations:

- Track the customer's evolving preferences and constraints across turns. Don't re-explain what's already been established.
- Build on prior context. If the customer narrows their interest, go deeper on that path rather than restating the full option space.
- When the customer changes direction or asks to reconsider, acknowledge their current position before exploring alternatives.
- If the customer is close to a decision, help them finalize with confidence -- address remaining concerns, suggest complementary items, or validate their choice.
\end{lstlisting}
\caption{System prompt used in the system prompt ablation. This prompt is prepended as the system message for all three frontier models in the ablation condition.}
\label{fig:sysprompt}
\end{figure*}

%% file: appendices/appendix_benchmark_variants.tex
\section{Benchmark Variants}
\label{app:benchmark_variants}
We release two variants of the benchmark:

\begin{itemize}
    \item \textbf{\benchnameFull{}} (525 missions, 10{,}863 rubrics): The complete benchmark comprising 232 single-turn queries and 293 multi-turn missions (1{,}764 turns) across five product families. This variant covers the full benchmark.


    \item \textbf{\benchnameHard{}} (108 missions, 1{,}663 rubrics): A subset of \benchnameFull{} containing missions where the nine-model average weighted pass rate falls below 60\%. This yields 69 single-turn queries and 39 multi-turn missions (304 turns), representing the missions that current models collectively struggle with. The Hard variant is designed for tracking progress on the most demanding shopping reasoning problems.
\end{itemize}

\begin{table}[tb]
\centering
\caption{Summary of the two \benchname{} variants.}
\label{tab:variants}
\small
\begin{tabular}{lccc}
\toprule
\textbf{Variant} & \textbf{Missions (ST / MT)} & \textbf{Turns} & \textbf{Rubrics} \\
\midrule
Full & 232 / 293 & 1{,}996 & 10{,}863 \\
Hard & 69 / 39 & 304 & 1{,}663 \\
\bottomrule
\end{tabular}
\end{table}

\paragraph{Selection criteria for \benchnameHard{}.}
For each mission, we compute the weighted pass rate (Eq.~\ref{eq:pass_rate}) per model, average across all nine evaluated models, and select missions where this average falls below 60\%. The threshold is applied uniformly across both splits. All five reasoning categories, all fifteen subcategories, all five product families, all six reasoning stages, and all six reasoning quality dimensions are represented in the Hard subset.

%% file: appendices/appendix_data_examples.tex
\section{Data Format and Examples}
\label{app:data_examples}

This appendix provides example benchmark data in the released JSON format.

\subsection{Single-Turn Example}

\paragraph{Example: Long-Haul Flight Headphones (st-10).}
\label{fig:data_example_st}
A single-turn mission with four rubrics (three required, one optional), all testing feature assessment.

\lstset{
  basicstyle=\scriptsize\ttfamily,
  breaklines=true,
  breakatwhitespace=true,
  breakindent=1em,
  columns=fullflexible,
  keepspaces=true,
  showstringspaces=false,
  upquote=true,
  xleftmargin=0pt,
  xrightmargin=0pt,
  frame=single,
  framerule=0.4pt,
  rulecolor=\color{csrbgray},
  aboveskip=6pt,
  belowskip=6pt,
  postbreak=\mbox{\textcolor{csrbgray}{$\hookrightarrow$}\space},
}
\begin{lstlisting}
{
  "mission_id": "st-10",
  "mission_name": "Long-Haul Flight Headphones",
  "mission_type": "Find Specific Solution",
  "mission_objective": "Customer is shopping for comfortable, long-battery-life,
    noise-canceling headphones for a 14 hour flight.",
  "product_family": "Hardlines",
  "time_sensitive": "Yes",
  "shopping_funnel_flow": ["Discover"],
  "turns": [
    {
      "reasoning_category": "Product Recommendation",
      "reasoning_subcategory": "Constrained Recommendation",
      "shopping_funnel_stage": "Discover",
      "messages": [
        {"role": "user", "content": "I need a pair of headphones for a 14 hour flight"}
      ],
      "rubrics": [
        {
          "text": "Discuss headphones that are comfortable enough for the customer
            to wear for 10+ hours.",
          "scope": "instance",
          "importance": "required",
          "reasoning_stage": "feature_assessment",
          "reasoning_quality": "relevance"
        },
        {
          "text": "Discuss headphones with battery life that will last 10+ hours
            with noise cancelation and have quick charging.",
          "scope": "instance",
          "importance": "required",
          "reasoning_stage": "feature_assessment",
          "reasoning_quality": "relevance"
        },
        {
          "text": "Discuss headphones that can cancel out chatter and engine noises.",
          "scope": "instance",
          "importance": "required",
          "reasoning_stage": "feature_assessment",
          "reasoning_quality": "relevance"
        },
        {
          "text": "Discuss portable headphones or portability features on headphones.",
          "scope": "instance",
          "importance": "optional",
          "reasoning_stage": "feature_assessment",
          "reasoning_quality": "relevance"
        }
      ]
    }
  ]
}
\end{lstlisting}

\subsection{Multi-Turn Example}

\paragraph{Example: Chocolate Making (mt-91, first 2 of 4 turns).}
\label{fig:data_example_mt}
A multi-turn mission progressing from Shopping Guidance to Product Recommendation across turns. Each turn carries independent rubrics with taxonomy tags.

\lstset{
  basicstyle=\scriptsize\ttfamily,
  breaklines=true,
  breakatwhitespace=true,
  breakindent=1em,
  columns=fullflexible,
  keepspaces=true,
  showstringspaces=false,
  upquote=true,
  xleftmargin=0pt,
  xrightmargin=0pt,
  frame=single,
  framerule=0.4pt,
  rulecolor=\color{csrbgray},
  aboveskip=6pt,
  belowskip=6pt,
  postbreak=\mbox{\textcolor{csrbgray}{$\hookrightarrow$}\space},
}
\begin{lstlisting}
{
  "mission_id": "mt-91",
  "mission_name": "how to make my own chocolate",
  "mission_type": "Explore & Discover",
  "mission_objective": "Customer is shopping for essential tools and supplies to
    begin making filled chocolates at home, seeking beginner-friendly
    chocolate-making equipment.",
  "product_family": "Consumables",
  "time_sensitive": "No",
  "shopping_funnel_flow": ["Discover", "Discover", "Explore", "Explore"],
  "turns": [
    {
      "reasoning_category": "Shopping Guidance",
      "reasoning_subcategory": "Decision-Factor Guidance",
      "shopping_funnel_stage": "Discover",
      "messages": [
        {"role": "user", "content": "I want to learn how to make my own
          chocolates, I want them to be able to have some sort of filling in it,
          what are some items that can help me get started?"}
      ],
      "rubrics": [
        {
          "text": "List at least 5 essential items for making filled chocolates.",
          "scope": "instance",
          "importance": "required",
          "reasoning_stage": "option_generation",
          "reasoning_quality": "completeness"
        },
        {
          "text": "For each item, briefly explain why it's necessary for making
            filled chocolates.",
          "scope": "instance",
          "importance": "required",
          "reasoning_stage": "domain_expertise",
          "reasoning_quality": "insightfulness"
        },
        {
          "text": "Recommend specific types of chocolate (e.g., couverture, candy
            melts) suitable for molding, explaining why they are preferred over
            regular chocolate chips.",
          "scope": "instance",
          "importance": "required",
          "reasoning_stage": "domain_expertise",
          "reasoning_quality": "insightfulness"
        },
        {
          "text": "Provide a brief explanation of tempering if mentioning
            couverture chocolate.",
          "scope": "instance",
          "importance": "optional",
          "reasoning_stage": "domain_expertise",
          "reasoning_quality": "insightfulness"
        },
        {
          "text": "Include a mention of a candy thermometer as an essential item.",
          "scope": "instance",
          "importance": "required",
          "reasoning_stage": "option_generation",
          "reasoning_quality": "completeness"
        }
      ]
    },
    {
      "reasoning_category": "Shopping Guidance",
      "reasoning_subcategory": "Decision-Factor Guidance",
      "shopping_funnel_stage": "Discover",
      "messages": [
        {"role": "user", "content": "what kind of filling is more beginner
          friendly, in terms of time and effort?"}
      ],
      "rubrics": [
        {
          "text": "Identify five beginner-friendly chocolate filling types.",
          "scope": "instance",
          "importance": "required",
          "reasoning_stage": "option_generation",
          "reasoning_quality": "concreteness"
        },
        {
          "text": "Explain why each recommended filling is suitable for beginners,
            focusing on ease of preparation, time, and effort.",
          "scope": "instance",
          "importance": "required",
          "reasoning_stage": "trade_offs",
          "reasoning_quality": "insightfulness"
        },
        {
          "text": "Provide concrete examples of how to flavor these fillings.",
          "scope": "instance",
          "importance": "required",
          "reasoning_stage": "actionability",
          "reasoning_quality": "concreteness"
        },
        {
          "text": "Do not recommend fillings that require complex techniques or
            many ingredients.",
          "scope": "instance",
          "importance": "required",
          "reasoning_stage": "option_generation",
          "reasoning_quality": "relevance"
        }
      ]
    }
  ]
}
\end{lstlisting}